\documentclass[sigconf]{aamas} 

\usepackage{balance} % for balancing columns on the final page

\setcopyright{ifaamas}
\acmConference[AAMAS '24]{Proc.\@ of the 23rd International Conference
on Autonomous Agents and Multiagent Systems (AAMAS 2024)}{May 6 -- 10, 2024}
{Auckland, New Zealand}{N.~Alechina, V.~Dignum, M.~Dastani, J.S.~Sichman (eds.)}
\copyrightyear{2024}
\acmYear{2024}
\acmDOI{}
\acmPrice{}
\acmISBN{}

\title[A Computational Framework for Human Values]{A Computational Framework for Human Values}

\author{Nardine Osman}
\affiliation{
  \institution{Artificial Intelligence Research Institute (IIIA-CSIC)}
  \city{Barcelona}
  \country{Catalonia, Spain}}
\email{nardine@iiia.csic.es}

\author{Mark d'Inverno}
\affiliation{
  \institution{Goldsmiths, University of London, UK}
  \city{}
  \country{}}
\affiliation{
  \institution{Artificial Intelligence Research Institute (IIIA-CSIC)}
  \city{}
  \country{}}
\email{dinverno@gold.ac.uk}

\begin{abstract}
There is an increasing recognition of the need to engineer AI that respects and embodies human values. The value alignment problem, which identifies that need, has led to a growing body of research that investigates value learning, the aggregation of individual values into the values of groups, the alignment of norms with values,  and the design of other computational mechanisms that reason over values in general. 
Yet despite these efforts, no foundational, computational model of human values has been proposed. 
In response, we propose a model for the computational representation of human values that 
builds upon a sustained body of research from social psychology. 
\end{abstract}


\newcommand{\BibTeX}{\rm B\kern-.05em{\sc i\kern-.025em b}\kern-.08em\TeX}

\usepackage{algorithm}
\usepackage{algpseudocode}
\algdef{SE}[DOWHILE]{Do}{doWhile}{\algorithmicdo}[1]{\algorithmicwhile\ #1}%

\usepackage{subcaption}

\newtheorem{definition}{Definition}
\newtheorem{property}{Property}

\newcommand{\eqdef}{\overset{def}{=}}  %NAR

 \usepackage{censor,caption}
\makeatletter
\gdef\@copyrightpermission{
	\begin{minipage}{0.3\columnwidth}
		\href{https://creativecommons.org/licenses/by/4.0/}{\includegraphics[width=0.90\textwidth]{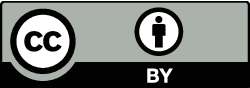}}
	\end{minipage}\hfill
	\begin{minipage}{0.7\columnwidth}
		\href{https://creativecommons.org/licenses/by/4.0/}{This work is licensed under a Creative Commons Attribution International 4.0 License.}
	\end{minipage}
	\vspace{5pt}
}
\makeatother

\begin{document}
\emergencystretch 3em 
%%% The following commands remove the headers in your paper. For final 
%%% papers, these will be inserted during the pagination process.

\pagestyle{fancy}
\fancyhead{}

%%% The next command prints the information defined in the preamble.

\maketitle 

%%%%%%%%%%%%%%%%%%%%%%%%%%%%%%%%%%%%%%%%%%%%%%%%%%%%%%%%%%%%%%%%%%%%%%%%

\section{Introduction}
Across governments, industry and the general public, there is an increasing recognition of the urgency for ethical approaches to AI (as evidenced by the numerous publications of ethics guidelines, e.g.~\cite{euArtificialIntelligenceAct,ec2019ethics,ieee,UNESCO,jobin2019global}). 
In academia, a growing body of research investigates the role of human values in designing ethical AI~\cite{russell2019human,GFAIH21,10.1007/s11023-020-09539-2,Weide2010}. 
Indeed, one of our leading AI research luminaries, Stuart Russell, believes the overarching goal of AI should change from ``intelligence" to ``intelligence provably aligned with human values"~\cite{russell2019human}.
This call to arms gave birth to the \emph{value alignment problem}.

This challenge of engineering values into AI in response to the value alignment problem has resulted in a range of research areas: how AI learns human values~\cite{Liu2019PersonalityOV,Lin2018AcquiringBK,89b26068890444aa88b4a15afe36625c,10.5555/3463952.3464048}; how individual values can be aggregated to the level of groups~\cite{Lera-LeriBSLR22}; how arguments that explicitly reference values can be made~\cite{Bench-Capon2009}; how decision making can be value-driven~\cite{TostoD12,ChhogyalNGD19,ijcai2017p26}; how online institutions can ensure value-aligned behaviours in hybrid communities~\cite{noriega23,noriega22}; and how norms are selected to maximise value-alignment~\cite{SerramiaLR20,MontesS21,abs-2110-09240}. 

% \paragraph{Problem and Proposed Solution.} 
Yet despite these efforts, no formal model of values exists that provides a concrete foundational platform from which data structures and algorithms can be designed to build AI architectures that address the value-alignment problem. 
In response, we present such a model and set out the following guiding principles: (i) employ a formal language to be precise and a foundation for proof and algorithmic development (e.g.~\cite{luck95,dinverno97}) (ii) ensure that our formal components lend themselves to data structure and algorithmic design (iii) subsume established concepts in AI research as much as possible, and (iv) make every effort to draw upon the wealth of work from within the humanities and social sciences (especially social psychology~\cite{Rohan2000}). 
With this in hand, we have a better opportunity to address Russell's challenge of moving from ``intelligence" to ``intelligence aligned with human values" by supporting agreement on foundational and explicit models and properties of values.

Our model is presented in four subsections in Section~\ref{sec:valueTax}.
The first is a formal definition of values and value taxonomies. 
The second models the values of individual agents and groups of agents (both artificial and human), such as with many online communities.  
In the third, we extend the model to incorporate the changing values of individuals and communities over time.
Finally, we address the problem of ensuring the extent (or degree) to which the behaviours of individuals or communities 
% (MAS) 
are aligned with an agreed set of values; the \emph{value alignment problem}. 
Our ongoing work with medical doctors at Hospital del Mar, Barcelona, is making use of our proposal to interrogate and model their four overarching bio-ethical values~\cite{Beauchamp1979-BEAPOB} (beneficence, non-maleficence, autonomy, and justice) and so develop AI that provides feedback on the alignment of medical decisions leading to better value-aware decision-making in the hospital. 

Each of those four subsections is divided into three parts: our proposal, a discussion of implementation choices, and a running example that embodies a concrete implementation choice. 
The running example should support the reader in understanding the development of our foundational proposal.
We then close with a reflection on the contributions and further work (Section~\ref{sec:StrengthLimit}) along with some concluding remarks (Section~\ref{sec:conclusion}).

%%%%%%%%%%%%%%%%%%%%%%%%%%%%%%%%%%%%%%%%%%%%%%%%%%%%%%%%%%%%%%%%%%%%%%%%****************

\section{A Formal Model for Value Representation}\label{sec:valueTax}

Our model for value representation provides the foundations for new computational mechanisms that reason over values, enabling artificial systems to make value-aware decisions explicitly. 

\subsection{What are values? A computational approach
% to value representation
}\label{sec:what}
Our approach ---aligned with Schwartz's Theory of Basic Values~\cite{Schwartz2012AnOO}--- views values as human abstract concepts that guide behaviour and whose exact meaning and interpretation vary both with the current context and over time. 
However, a concrete computational representation of values requires a machine to reason with values. 
For example, while we might talk about fairness in general, for a specific application supporting a mutual aid community, fairness might be understood as ``one does not ask for help more than what they volunteer.''. That is, the abstract concept of fairness acquires a concrete definition (grounding semantics) through a {\it property} whose satisfaction (or degree of satisfaction) can be automatically verified in a given system (more details are provided in Subsection~\ref{sec:ex-vtax}). In this case, the property specifies that you cannot ask for more help than you volunteer. 

\begin{figure}
\centering
\includegraphics[width=\linewidth]{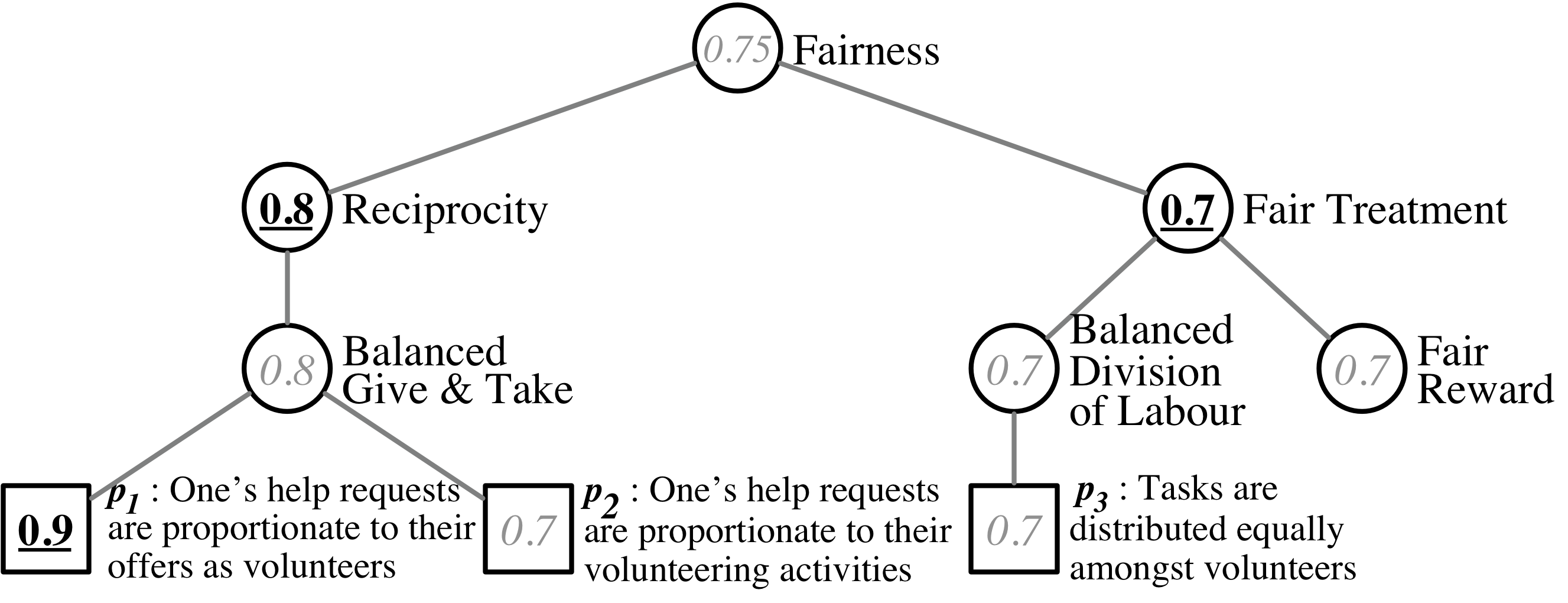}
\caption{Abstractions for the value fairness: 
numbers indicate node importance, specified (black) and deduced (gray)}
\label{fig:egGeneral}
\end{figure}

Of course, there may be different levels of abstraction and grounding semantics for a value. For example, say the application was to be adopted by another community where volunteers explicitly support older people. The new community might find the old view of fairness ---that one does not ask for help more than what they volunteer--- unsuitable because this community expects to have volunteers that support older people, and they usually only ask for help with their day-to-day tasks without volunteering. As such, this new community should state that fairness is not about balanced give and take but about balanced workload division amongst the volunteers. Here, we encounter different levels of abstraction and grounding properties, illustrated in Figure~\ref{fig:egGeneral}. 

The top node
%in Figure~\ref{fig:egGeneral} 
presents the abstract concept of fairness; the middle nodes present different concepts of fairness that are more specific than the top node, highlighting different abstraction levels and interpretations; and the bottom nodes present properties that ground the semantics of abstract concepts and allow for a computational evaluation of values. 
We use square nodes for nodes that ground the semantics of abstract concepts.  

This interplay between the abstract and the grounding semantics of values is reflected through the relations between nodes, as shown in Figure~\ref{fig:egGeneral}. 
Different abstraction levels for value concepts may exist, along with different grounding semantics enabling computational approaches over values. 
Another key concept from Schwartz is \emph{value priority}, which determines how \emph{important} a value is for individuals or groups. In other words, it is not just the semantics of fairness that influences the behaviour of an individual or a group, but how important fairness as a value is for that individual or group. 
We model \emph{value importance} by attaching a measure of \emph{importance} to each node of Figure~\ref{fig:egGeneral}.
We propose that values are computationally modelled as taxonomies, where a high-level value concept becomes more specific as one travels down any taxonomy path, becoming concrete and computational at leaf nodes. 
This is consistent with research in value-sensitive design~\cite{vandePoel2018} on value change taxonomies, as well as with Schwartz and a wide range of research from social psychology~\cite{Schwartz2012AnOO,Rohan2000}.

Furthermore, using taxonomies allows for easy navigation between abstract and concrete grounding semantics of values, which supports humans in understanding values and reflecting on and deliberating about them. It may be argued that all a machine requires are the property nodes to ensure behaviour alignment with values (as illustrated in Subsection~\ref{sec:why}). However, in the case of having a new community emerging, such as the community of volunteers supporting older people, we saw how a new community might have its local and distinct view of the fairness value. 
A taxonomy representation makes the deliberation process on the different semantics of fairness and the evolution of these semantics over time and context possible. 
We expect semantics to continuously evolve with the requirements of new interactions, as detailed in Subsection~\ref{sec:context}. In addition, we believe a taxonomy can support the deliberation process over values and their importance. 
Definition~\ref{def:valueTaxonomy} presents our proposed taxonomy-based value representation.

\begin{definition}[Value taxonomy]\label{def:valueTaxonomy}
A value taxonomy $\mathcal{V}=(N,E,I)$ is defined as a directed acyclic graph, where: 
\begin{enumerate}
    \item The set of nodes $N=N_{l}\cup N_{\phi}$ represents value concepts, and it is composed of two types of nodes: i) those that are specified through labels, with $N_{l} \subset \mathrm{L}$ representing the set of label nodes and $\mathrm{L}$ is the set of all value labels representing abstract value concepts like `fairness' or `reciprocity'; and ii) those that are specified through concrete properties, with $N_{\phi} \subset \Phi$ representing the set of property nodes and $\Phi$ is the set of all value properties whose satisfaction can be automatically verified at different world states, such as having the number of times one offers help in a mutual aid community larger than the number of times one asks for help. 
    \item The set of edges $E: N \times N $ is a set of directed edges $(n_{p},n_{c}) \in E$ that represent the relation between value concepts $n_{p}$ and $n_{c}$ (the parent and child nodes, respectively) illustrating that the value concept $n_{p}$ is a more general concept than $n_{c}$. 
    \item The importance function $I: N \to COD_{\bot}$ either assigns an importance measure from the codomain $COD$ to value concepts in $N$, or assigns $\bot$ to value concepts when their importance measure is undefined.
\end{enumerate}
\end{definition} 

\begin{figure}[!hb]
\centering
\includegraphics[width=0.25\textwidth]{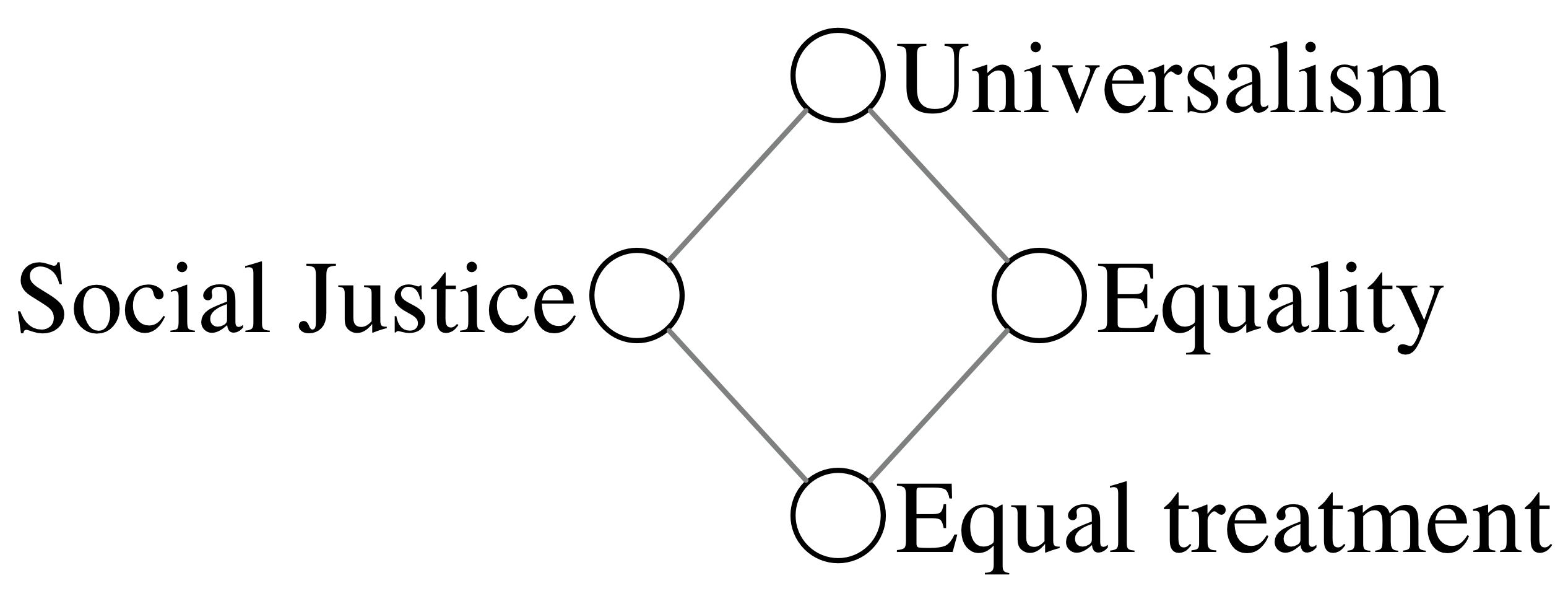}
\caption{Value concepts with more than one parent node}
\label{fig:multiple_parent_nodes}
\end{figure}

Note that we specify the value taxonomy as a directed acyclic graph instead of the more traditional taxonomy tree because one value concept (node) may have more than one parent node. For example, the value `equal treatment' in the taxonomy of Figure~\ref{fig:multiple_parent_nodes} can act as a more specific concept for both the `social justice' and `equality' values (where the social justice and equality values and their being examples of the universalism value are taken from Schwartz' value hierarchies~\cite{Schwartz2012AnOO}).

We also require property nodes to be restricted to leaf nodes. In other words, the concrete grounding semantics of a value concept (specified as a property node) cannot be more general than another value concept (specified as another property or label node). For example, in Figure~\ref{fig:egGeneral}, the property of having one's help requests proportionate to their volunteering offers cannot be more general than abstract concepts such as balanced give and take. This requirement is defined formally as follows:
     $\nexists \; (n_{p},n_{c}) \in E \cdot n_{p} \in N_{\phi}$ .
We introduce this condition to simplify the construction and interpretation of value taxonomies. 
Future work may relax this requirement if property subsumption mechanisms are introduced to ensure parent nodes subsume children nodes. 

There needs to be coherence regarding value importance within a taxonomy. 
Whilst humans are not always coherent, AI systems can be, and so can support humans to be more so. 
We express what we mean by coherence by requiring that the importance of a parent node be consistent with the combined importance of all of its child nodes. 
For example, if the importance of all children nodes is low, then the importance of the parent node cannot be high, and vice versa. We formally define the coherence of value importance accordingly:
\begin{definition}[Coherence of value importance]\label{def:coherence} 
Importance within a value taxonomy $\mathcal{V}=(N,E,I)$ is said to be coherent if and only if, for all nodes $n\in N$ with children nodes (there exists $(n,n_{c})\in E$), the importance of the parent node is an aggregation of the importance of its children nodes:
    \begin{equation}\label{eq:coherence}
        \forall \, n\in \{n_{p}\,|\, (n_{p},n_{c}) \in E\} \;\cdot\; I(n) = \mathop{\mathbf{A}}\limits_{n'\in X_{n}} I(n')
    \end{equation}
where $X_{n} = \{n_{c} \,|\, (n,n_{c}) \in E \}$ is the set of all children nodes of $n$, and $\mathbf{A}: COD^{m} \to COD$ is an aggregation function that takes a set of size $m\in \mathbb{N}^{*}$ of importance measures in $COD$ (specified as $COD^{m}$) and returns an aggregation of those measures, where the aggregation also falls in the same range $COD$.
\end{definition}

The importance measures of some of the taxonomy nodes might either be provided manually by humans or learned from other sources, such as past interactions. 
A coherence mechanism is then needed to ensure importance measures are coherent. Then, propagation mechanisms could be constructed to calculate the importance of nodes that have not been provided, building on existing propagation mechanisms~\cite{OsmanSS10,LiquidPub}. Any propagation mechanism must respect the coherence of value importance within the taxonomy. 

But what about the aggregation function $\mathbf{A}$? We argue that symmetry, idempotence and monotonicity are desirable properties to be held by any such function.
For instance, when calculating a parent node's importance, ordering its children nodes' importance measures should not affect the aggregation (symmetry). 
Also, if we assume all children nodes of some parent node to have the same importance $i$, then we believe the parent node should share that importance too (idempotence). It should neither be more important nor less important than $i$. 
Finally, increasing the importance measure of the children nodes should not decrease the parent node's importance (monotonicity). 
Formal definitions of these properties follow.

\begin{property}[Symmetry of aggregation]\label{prop:symmetry} 
The aggregation function $\mathbf{A}$ is symmetric if, for all sets of importance measures $\lambda \in COD^{m}$ and all permutations $\pi \in \Pi_{\lambda}$ of those sets (where $\Pi_{\lambda}$ is the set of all permutations of $\lambda$), we have: 
$\mathbf{A}(\lambda) = \mathbf{A}(\pi)$
\end{property}
That is, the order of aggregated values does not matter.

\begin{property}[Idempotence of aggregation]\label{prop:idempotence}
An aggregation function $\mathbf{A}$ is idempotent if, for all %$m\in \mathbb{N}^{*}$ and all $\lambda \in COD^{m}$ where $\lambda = \{i, \ldots, i\}$
%importance measures 
$i\in COD$, we have: 
$\mathbf{A}(i, \ldots, i) = i $
\end{property}
That is, the aggregation of several instances of the same measure will return that same measure.

\begin{property}[Monotonicity of aggregation]\label{prop:monotonicity} 
An aggregation function $\mathbf{A}$ is monotonous if, for all sets of importance measures $\lambda, \lambda' \in COD^{m}$, we have: 
$ (\forall \; 0<p\leq m \; \cdot \; \lambda_{p} \leq \lambda'_{p}) \;\; \Rightarrow \;\; \mathbf{A}(\lambda) \leq \mathbf{A}(\lambda')$ , 
where $\lambda_{p}$ represents the element in position $p$ of the set $\lambda$.
\end{property}
That is, if the measures in $\lambda$ are smaller than or equal to those in $\lambda'$ (per position), then the aggregation of the measures in $\lambda$ should be smaller than or equal to the aggregation of those in $\lambda'$. 
 
Our requirements for Properties \ref{prop:symmetry}--\ref{prop:monotonicity} help define the type of aggregation operator. %, as we show next.
First, from~\cite[p.~14]{12-Ma98}, we know that the idempotence and monotonicity properties imply compensativeness.
Compensative operators are aggregation operators that are neither conjunctive nor disjunctive. They are limited between the $\min$ and $\max$, which are the bounds of the t-norm and t-conorm families. This implies that for any set of importance measures (for all $\lambda \in COD^{m}$), the aggregation will fall between the minimum and maximum measures in that set:
$\min \lambda \leq \mathbf{A}(\lambda) \leq \max \lambda $.

We believe falling between the minimum and maximum is appropriate for our setting: a parent node should not be more important than its most important child node nor less important than its least important child node.
Furthermore, from~\cite[p.~13]{12-Ma98}, we also know that compensative aggregation operators that satisfy the symmetry and monotonicity properties are averaging operators.
As such, we propose $\mathbf{A}$ to be an averaging operator, though the exact choice of this operator is left for implementation.

\subsubsection{Implementation Choices}\label{sec:what-imp} 
\paragraph{Specifying property nodes.} 
We use properties to specify the grounding semantics of values, as properties have traditionally been used to describe the world state, and their satisfaction can be computed~\cite{Stirling2001}. 
The exact language used for specifying properties is an implementation decision. 
Our proposal is agnostic regarding any choice of implementation language or algorithmic design. 
It can encompass any theory written in propositional, first-order, deontic or modal logic.

We note that the use of properties may initially appear to embody a consequentialist view, where only the outcome of behaviour is what matters. 
However, in our model, properties can be designed to evaluate not only action outcomes but also attached to actions themselves (such as whether an action is permitted, e.g. lying is never acceptable). 
The expressiveness of the chosen language for implementing properties plays a significant role here.
In this paper, we limit the examples to propositional logic and, more specifically, simple propositions to improve readability. %Complex languages with higher expressivity can be chosen. 

\paragraph{Choosing the codomain of value importance.}
Concerning the importance measure of a value concept, the choice of the codomain $ COD $ that evaluates this importance $ I $ is also an implementation decision. Schwartz, for instance, used the range $\{-1,0,1,2,3,4,5,6,7\}$ for people to specify the importance of a value, such as `equality' (equal opportunity for all) or `pleasure' (gratification of desires), as a guiding principle in their lives~\cite{Schwartz2012AnOO}, where $-1$ represents opposing a value, $0$ represents considering the value to be non-important, and positive numbers represent
different degrees of supporting a value ---`supreme importance', `very important', `important', and so on. 

Alternative approaches, such as defining importance as a partial or total order, may also be considered. Schwartz argues that this order (the relevant importance~\cite{Schwartz2012AnOO}) should be used when reasoning about values, where the order is deduced from the importance assigned by people. 
A partial order may sometimes be more intuitive for humans to provide, as opposed to giving numerical importance measures. When this is the case, human stakeholders may use partial orders to specify importance measures and conversion mechanisms may be used to translate partial orders into numeric values. 
In general, the implementation choice must depend on the specific application's requirements.
In this paper's example, we set the codomain to $[-1,1]$ as reasoning with numbers is computationally easier than reasoning with partial orders. Furthermore, the range $[-1,0]$ is more intuitive for describing opposition to a value. This expands Schwartz's opposition measures, as there may also be degrees when opposing a value the same way there are degrees in supporting a value. 

\paragraph{Ensuring the coherence of value importance.} 
As for the aggregation function $\mathbf{A}$ that ensures the coherence of value importance within a taxonomy, different averaging operators may be investigated. In this paper, we propose a simple average:
\begin{equation}\label{eq:coherence_imp}
    I(n_{p}) = (\displaystyle\sum_{n_{c}\in X_{n_{p}}} I(n_{c})) \;/\; (|X_{n_{p}}|)
\end{equation}

Recall that we assume the importance measure of some nodes to either be provided manually or learnt. A propagation mechanism may then be implemented to calculate the importance of nodes that have not been manually provided or learnt, all while ensuring the overall coherence of value importance within the taxonomy (Definition~\ref{def:coherence}).
An example propagation mechanism can be found in~\cite[Alg. 1]{ValuesFrameworkArXiV23}.

\subsubsection{The Running uHelp Example}\label{sec:ex-vtax}

We'll consider the value {\it fairness} for our running example. This value has emerged as an important value in the participatory design meetings 
with potential users of the uHelp app.\footnote{The uHelp app (\texttt{https://uhelpapp.com}) was developed at IIIA-CSIC in collaboration with the first author of this article. It is available on Google Play (\texttt{https://play.google.com/store/apps/details?id=es.csic.iiia.uhelpapp}) and Apple Store (\texttt{https://itunes.apple.com/es/app/uhelp/id1089461370}).}
The uHelp app is a social networking app that allows people to find help within their social network with their day-to-day activities, such as finding someone to substitute for them at work tomorrow or someone to lend them some chairs for a party~\cite{KosterMOSSSFJPG13,KosterMOSSSJFPG12,uhelpARXIV}. 
We choose not to confine our model to predefined value systems, like Schwartz's infamous set of universal human values~\cite{Schwartz2012AnOO}, but opt for values that may emerge from different stakeholders. 

Recognising relevant high-level values (root nodes) may be achieved either through a bottom-up approach, in which relevant values emerge from discussing the context with the relevant stakeholders, or through a top-down approach, in which a given stakeholder specifies the relevant values from prior knowledge.
The choice is usually use case dependent. For example, in other interactions with doctors and firefighters, relevant values were predefined by the relevant institutions and provided in a top-down approach. 
Once this has taken place the rest of the taxonomy emerged from contemplating those values and how they are understood in a given context. As new contexts emerged over time, the taxonomy was updated to consider the requirements of these new contexts. For example, fairness was initially understood as balanced give and take for a mutual aid community of single mothers. In other words, the users wanted to avoid greedy members asking for help and never offering it. But when, later on, we discussed uHelp with a community providing support to older people, it became clear that a balanced give-and-take was against their values.
A new understanding of fairness emerged: the balanced division of labour. In other words, while some community members will ask for help and others will provide it, the workload among those giving help should be balanced. 
Figure~\ref{fig:egGeneral} illustrates the evolving value taxonomy for fairness from uHelp's perspective. 
Next, we define some of the grounding semantics of those abstract concepts, i.e. defining the property leaf nodes. Recall that property nodes are presented as squares, whereas label nodes are shown as circles.

%\begin{figure}
%    \centering
%    \includegraphics[width=\linewidth]{figures/uHelp_fairness_taxonomy.png}
%    \caption{uHelp's value taxonomy for the value \emph{fairness}}
%    \label{fig:valueTaxonomy}
%\end{figure}

For illustrative purposes, we provide two different computational approaches defining balanced give and take using properties $p_1$ and $p_2$. The first states that one's help requests are proportionate to the number of times the user offered help, whereas the second states that one's help requests are proportionate to the number of times the user was chosen to help (because not all those offering help get selected). One straightforward approach to specifying $p_1$ and $p_2$ is through the ratio of requests to offers/volunteering being greater than 1, as illustrated in property definitions~\ref{eq:p1} and~\ref{eq:p2}. Of course, we provide simple properties for better readability, but we note that properties can get as complex as the language allows (remember that the choice of language for specifying properties is an implementation choice, and one may choose to work with highly expressive languages).
\begin{equation}\label{eq:p1}
    p_1 \quad \eqdef \quad  (\#_{requests}) \;/\; (\#_{\mathit{offers}}) > 1
\end{equation}
\begin{equation}\label{eq:p2}
    p_2 \quad \eqdef \quad  (\#_{requests}) \;/\; (\#_{volunteering}) > 1
\end{equation}
The two different properties for balanced give and take illustrate how different semantics may be provided for the same abstract value. In this specific case, one can imagine $p_2$ to have been initially defined, but after some interactions, some users are never chosen despite their volunteering, and as such, the system prevents them from asking for help so that balanced give and take is respected. To compensate for this, $p_1$ gets added to the taxonomy so that balanced give and take considers the offers for help instead of being chosen. This illustrates how the taxonomy may evolve by learning from experience what the best grounding semantics (property) of a given abstract concept are. 

The computational approach, or the property node, describes the balanced division of labour through property $p_3$, which states that tasks are equally distributed amongst volunteers. One approach to specifying $p_3$ is through the difference between the uniform distribution $U$ and the distribution $D$ of the numbers of tasks assigned to each volunteer, where this difference should be smaller than a predefined threshold $\epsilon$ (property definition~\ref{eq:p3}).
\begin{equation}\label{eq:p3}
    p_3 \quad \eqdef \quad  \mathit{difference}(D,U) < \epsilon
\end{equation}
where the difference between two distributions may be calculated using approaches like the Kullback–Leibler divergence~\cite{mackay2003information} or the earth mover's distance~\cite{emd}. 

This taxonomy does not specify the property node providing the grounding semantics for fair reward. Indeed, some abstract concepts may remain abstract for some time.
Furthermore, we note that while stakeholders discuss and agree on abstract value concepts, engineers must define the properties that ground the semantics of those values. Alternatively, AI may support constructing and updating these value taxonomies based on emerging value learning mechanisms~\cite{10.5555/3463952.3464048,VALE2023-SONY}.

A crucial feature of the value taxonomy is the importance of nodes. We imagine each context to have its own taxonomy with its own importance measures (see Subsection~\ref{sec:context}). As such, an adapted uHelp taxonomy for fairness will be made available for the community of single mothers and another for the community supporting older people. However, whether importance measures are available for the general taxonomy of uHelp is application-dependent.

Getting stakeholders to specify the importance of all nodes is usually challenging, even if importance is provided through a partial order. As such, we give an example where importance is assigned to a subset of these nodes (Figure~\ref{fig:egGeneral}), where we represent importance measures explicitly provided through bold underlined numbers. From those, our propagation mechanism (see~\cite[Alg. 1]{ValuesFrameworkArXiV23}) then calculates possible coherent importance measures for the remaining nodes (the non-underlined numbers).
We have provided the value taxonomy of one uHelp value: fairness. We imagine applications to have several relevant values and, thus, several such taxonomies.

\subsection{How do values change with context? Context-based value taxonomies}\label{sec:context}
Our stance is that values are context-dependent. This is the stance of value-sensitive design~\cite{vandePoel2018}, as well as the stance of many social psychologists before them~\cite{Rohan2000}.  
We argue that we all have a general view of what a value is, defined through its value taxonomy, and this view evolves with our experiences, where new nodes (label-based and property-based) are continuously added.
If a new context requires adding new nodes, those are added to the general taxonomy. This is how, in general, taxonomies evolve with experiences. 

Suppose a new context necessitates eliminating existing nodes. This is achieved by setting their importance to zero for that specific context (if zero is the neutral-based measure of the chosen codomain). Moving from a general taxonomy to a context-based one is implemented by simply revisiting the importance measures of the general taxonomy (whether they were defined or not), making some nodes or branches more prominent than others. 
We note that if property-based leaf nodes did not exist for a prominent branch, they must be added to the general taxonomy. 
Otherwise, a computational approach considering those branches will not be possible. 
Our definition of a context-based value taxonomy is presented next. 
\begin{definition}[Context-based value taxonomy]\label{def:contextValueTaxonomy}
A context-based taxonomy $\mathcal{V}_{c}=\{N,E,I_{c}\}$ is an alteration of a general taxonomy $\mathcal{V}=\{N,E,I\}$ where the importance of nodes are updated for the given context. 
The importance of nodes in the context-based taxonomy $\mathcal{V}_{c}$ is independent of the importance of those nodes in the general taxonomy $\mathcal{V}$. 
The function calculating the importance of nodes
% within this context-based taxonomy 
with respect to a context $c$ is defined as $I_{c}: N \to COD_{\bot}$, where $COD_{\bot}$ defines the union of $COD$ with the undefined variable ($\bot$). 
\end{definition}
%This function assesses the importance of nodes with respect to a context $c$, where the context is defined through a set of properties $P_{c} \subseteq P$. The context $c$ is considered to hold ($holds(c)$) ---that is, we can say that we are in that context--- when its properties are satisfied: $(\forall p\in P_{c} \, \cdot \models p ) \implies \models holds(c)$.
%
% This provides us with a way of defining a context: i.e., a set of properties $P_{c} \subseteq P$. The context $c$ is then considered to hold ($holds(c)$) ---that is, we can say that we are in that context--- when its properties are satisfied: $(\forall p\in P_{c} \; (\models p) ) \implies \models holds(c)$.

\subsubsection{Implementation Choices}\label{sec:how-imp} 
\paragraph{Constructing Context-Based Taxonomies.} 
Different mechanisms for evaluating the importance of nodes in context-based taxonomies may be developed. One possible implementation (see~\cite[Alg. 2]{ValuesFrameworkArXiV23}) would take a bottom-up approach where only the importance of property nodes for the given context is evaluated (regardless of how they are obtained, whether they are provided manually by stakeholders or learnt from similar past experiences of similar contexts). This approach aims to better assess the importance of concrete property nodes in specific contexts. Then, the importance of the remaining nodes in the taxonomy is calculated by propagating the importance measures of the property nodes across the taxonomy, as described in Subsection~\ref{sec:what-imp}. 

Alternative implementations may be investigated. Rather than starting with property nodes and following a bottom-up approach, a top-down approach may be implemented to assess abstract concepts regardless of their grounding semantics. Other domains suggest a mix. Finally, we note that inconsistencies may arise between general and context-based taxonomies. This is expected and entirely normal. For example, while reciprocity might be considered very important as an abstract concept for the uHelp app, it might be regarded as less important for a specific context, such as for the community of volunteers supporting older people (see Subsection~\ref{sec:howEG}).

\paragraph{Visualising Taxonomies.} 
In addition to constructing context-based taxonomies through updating importance measures and for improved visualisation, we propose removing parts of the taxonomy deemed irrelevant for a given context. This step is optional but may be useful when visualising growing taxonomies for a given context. This also helps stakeholders quickly spot the relevant nodes for this context.

Different approaches may be followed when deciding which branches are relevant %, and as such, 
hiding irrelevant details. For example, one approach (see~\cite[Alg. 2]{ValuesFrameworkArXiV23}) is to maintain the branches that lead to relevant property nodes and eliminate those that lead to irrelevant property nodes, where the importance of nodes dictates. Relevance may be domain dependent and so determined by the chosen codomain. In this implementation of our model, irrelevant nodes are those with zero importance.

\subsubsection{The Running uHelp Example}\label{sec:howEG} 
\begin{figure*}[!t]
    \centering
    \begin{subfigure}[b]{.32\linewidth}
         \centering
         \includegraphics[width=\linewidth]{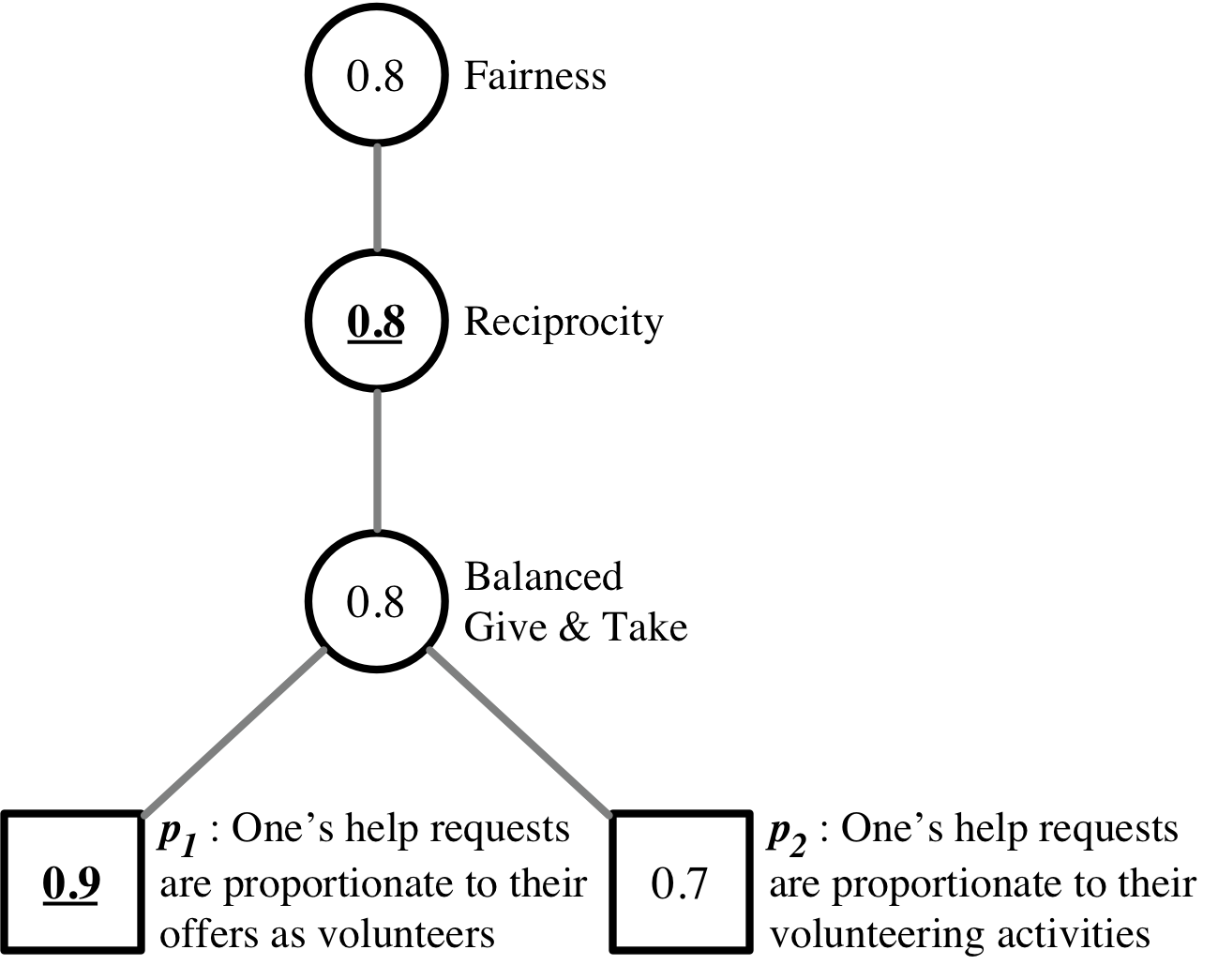}
         \caption{$\mathcal{V}_{c_{s}}$ for the community of single mothers, which happens to be the same as the initial general uHelp taxonomy $\mathcal{V}$}
         \label{fig:single-mothers-1}
     \end{subfigure}
     \hfill
     \begin{subfigure}[b]{.32\linewidth}
         \centering
         \includegraphics[width=\linewidth]{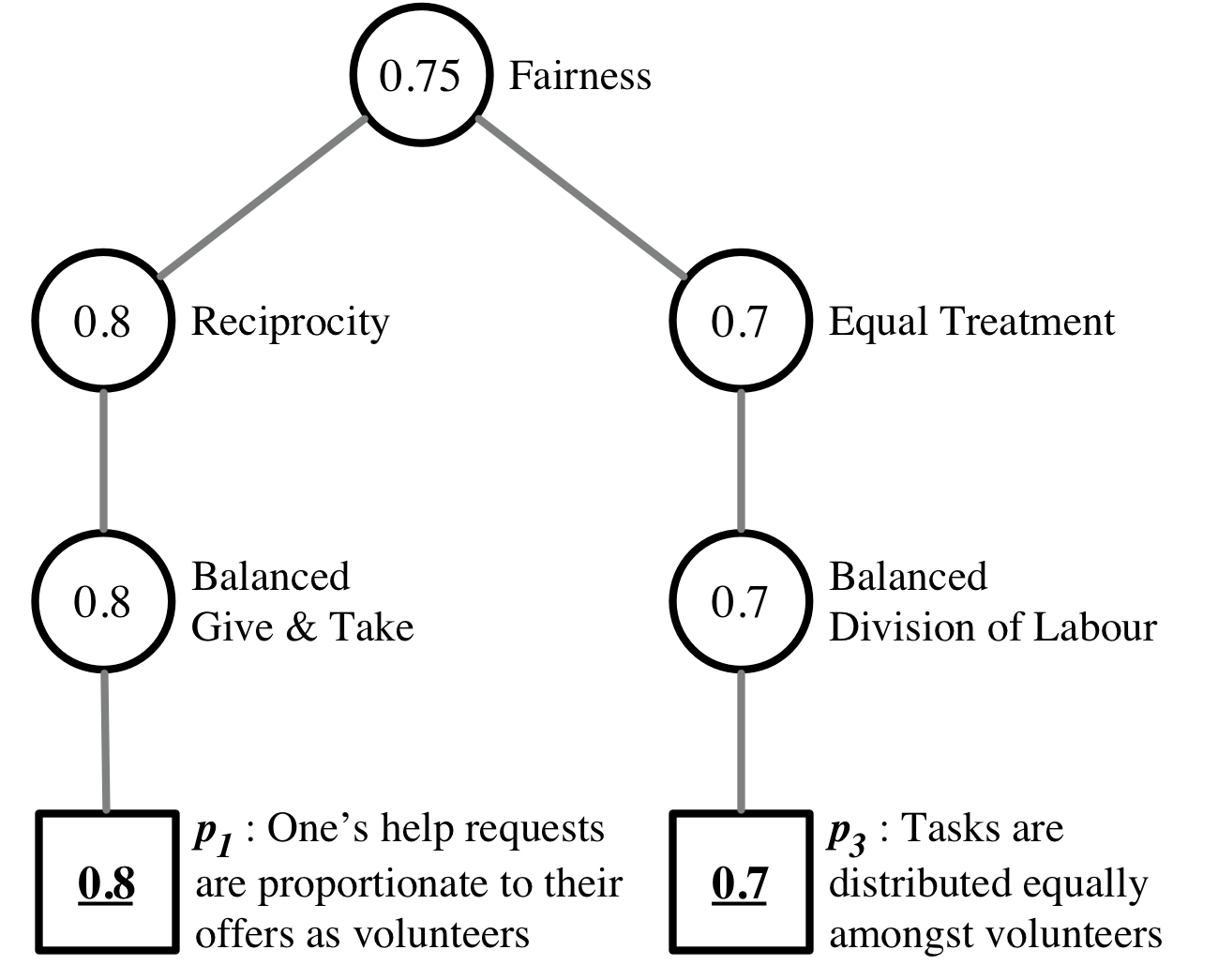}
         \caption{$\mathcal{V}_{c_{s}}'$ for the updated context-based taxonomy for the community of single mothers\\}
         \label{fig:single-mothers-2}
     \end{subfigure}
     \hfill
     \begin{subfigure}[b]{.32\linewidth}
         \centering
         \includegraphics[width=\linewidth]{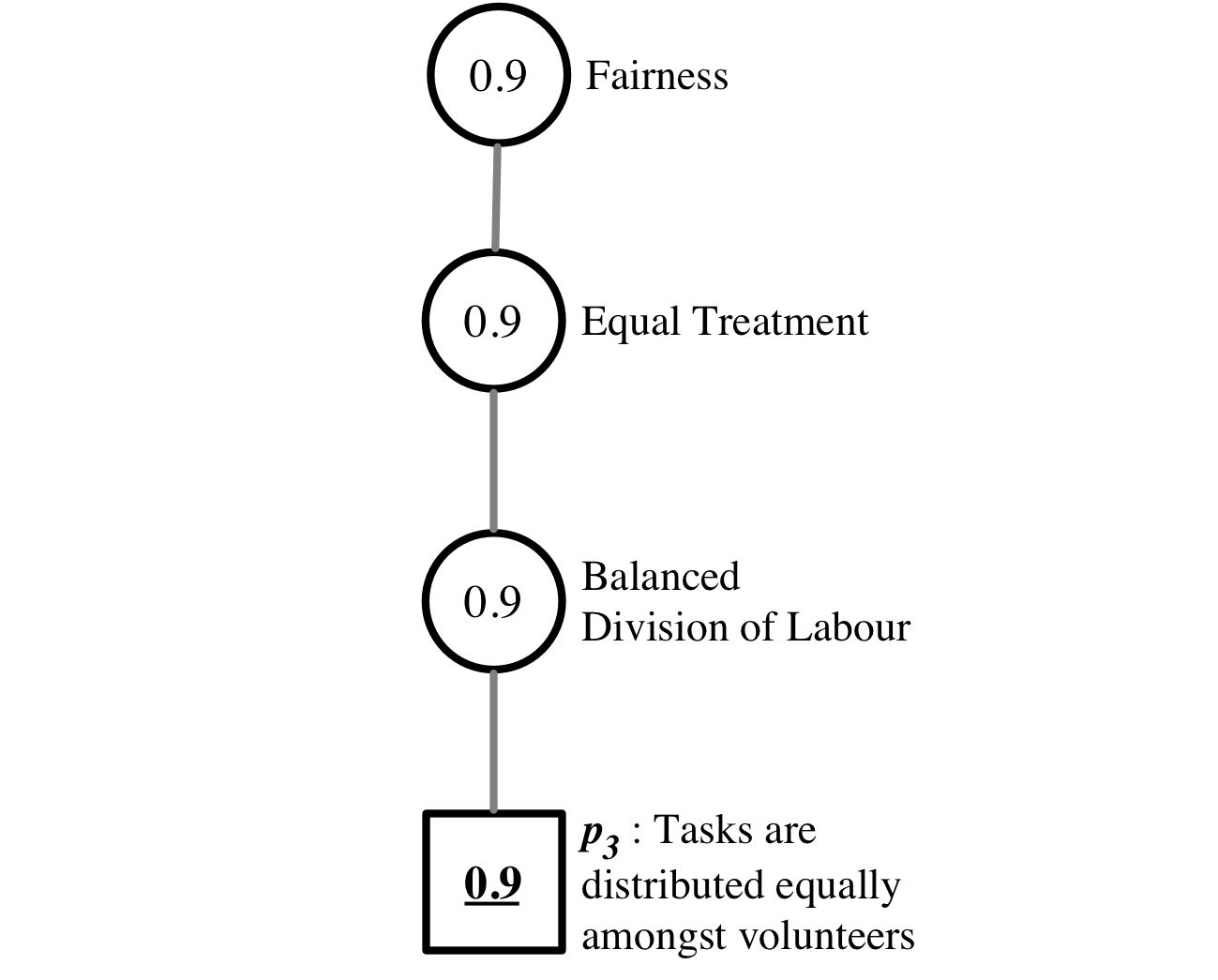}
         \caption{$\mathcal{V}_{c_{e}}$ for a community of volunteers supporting the older people\\}
         \label{fig:elderly}
     \end{subfigure}
    \caption{Different context-based value taxonomies for fairness in uHelp}
    \label{fig:contextValueTaxonomy}
\end{figure*}

While Figure~\ref{fig:egGeneral} has presented uHelp's general taxonomy for fairness, we present examples of context-based taxonomies for different uHelp communities in this subsection. As illustrated in Subsection~\ref{sec:ex-vtax}, the first community to adopt uHelp was the community of single mothers (whose context is specified as $c_{s}$). For them, fairness implied a balanced give and take. We could imagine that at the beginning, both the general taxonomy and the context-based taxonomy for the $c_{s}$ community were defined through the taxonomy of Figure~\ref{fig:single-mothers-1}. Then, as another context emerges, which is that of the community supporting older people (whose context is specified as $c_{e}$), new nodes were added to the general taxonomy, resulting in the Figure~\ref{fig:egGeneral} taxonomy. 

Armed with this new general taxonomy for fairness, the community supporting older people decides that the requirement of having a proportionate number of help requests with respect to the number of help offers ($p_1$) or volunteering ($p_2$) is irrelevant. In contrast, the requirement that labour should be divided in a balanced way amongst all volunteers is of utmost importance. 
So the importance of the property nodes of  Figure~\ref{fig:egGeneral} are specified as follows: 
%
%First, we note that according to Definition~\ref{def:contextValueTaxonomy}, $I_{c_{e}}(p_{1},P_{c_{e}})$ represents the importance of node $p_{1}$ in the context of the community supporting older people, where that context is defined through the properties $P_{c_{e}}$. In the rest of this paper, we simplify notation by assuming $I_{c}$ for the case when the properties $P_{c_{e}}$ hold, as follows: $I_{c_{e}}(p_{1})$.
%
% Using this notation 
%So we have:
%
$$\begin{array}{ccccc}
I_{c_{e}}(p_{1}) = 0 & ; &
I_{c_{e}}(p_{2}) = 0 & ; &
I_{c_{e}}(p_{3}) = 0.9 %\phantom{-}0.9
\end{array}$$

The context-based taxonomy of this new community is visualised (in our approach) by eliminating all branches leading to property nodes with zero importance, leaving us with one single branch leading to property node $p_3$. The resulting taxonomy is presented in Figure~\ref{fig:elderly}. The importance of each of the upper nodes inherits the importance of the property node, following our proposed propagation mechanism (see~\cite[Alg. 1]{ValuesFrameworkArXiV23}). 
With this new general taxonomy (Figure~\ref{fig:egGeneral}), the single mother's community decides to revise their context-based taxonomy. They like the new grounding semantics that have emerged and determine that it is important that requests should be both proportionate to offers ($p_1$), and equally distributed amongst volunteers ($p_3$). 
The new importance then reflects this new view measures they assign to the property nodes of Figure~\ref{fig:egGeneral}:
$$\begin{array}{ccccc}
I_{c}(p_{1}) = 0.8 & ; &
I_{c}(p_{2}) = 0\phantom{.0} & ; &
I_{c}(p_{3}) = 0.7
\end{array}$$

Their new taxonomy is updated and visualised by eliminating branches that lead to property nodes with zero importance. This results in Figure~\ref{fig:single-mothers-2}. 
As before, the importance of the upper nodes is inherited from the importance of the property nodes.

\subsection{Who holds values? Individuals vs collectives}\label{sec:who}
Entities hold values: values do not exist on their own. In other words, there is no universal value taxonomy for fairness. Different people or groups will hold different views on values, leading to different taxonomies. We use the notation $\mathcal{V}^{x}$ to represent the value taxonomy held by entity $x$, where $x$ may represent an individual $i_j$ or a collective $\{i_1,\ldots, i_n\}$. We use the word collective to describe a group of interacting individuals,
which may be a community, an organisation, an institute, a society, a culture, etc. When a collective holds a value taxonomy, this is understood as the value taxonomy describing the values of the collective as a whole and not its individuals (or interacting members). 
Individuals may not all have their value taxonomy aligned with the collective's. The issue of how the collective agrees on its taxonomy is left for future work.
To simplify notation, in the rest of the paper, we drop the $x$ from $\mathcal{V}^{x}$ when it is clear who holds the taxonomy.
As autonomous agents, humans (and future software agents) do not only understand their own values or the values of the collectives they belong to, but they also observe others. They can form beliefs about the values of others. We use the notation $\mathcal{V}^{x>y}$ to represent what $x$ believes are the values of $y$, where both $x$ and $y$ may represent an individual or a collective. 

\begin{figure*}[!t]
    \centering
    \begin{subfigure}[b]{.24\linewidth}
         \centering
         \includegraphics[width=\linewidth]{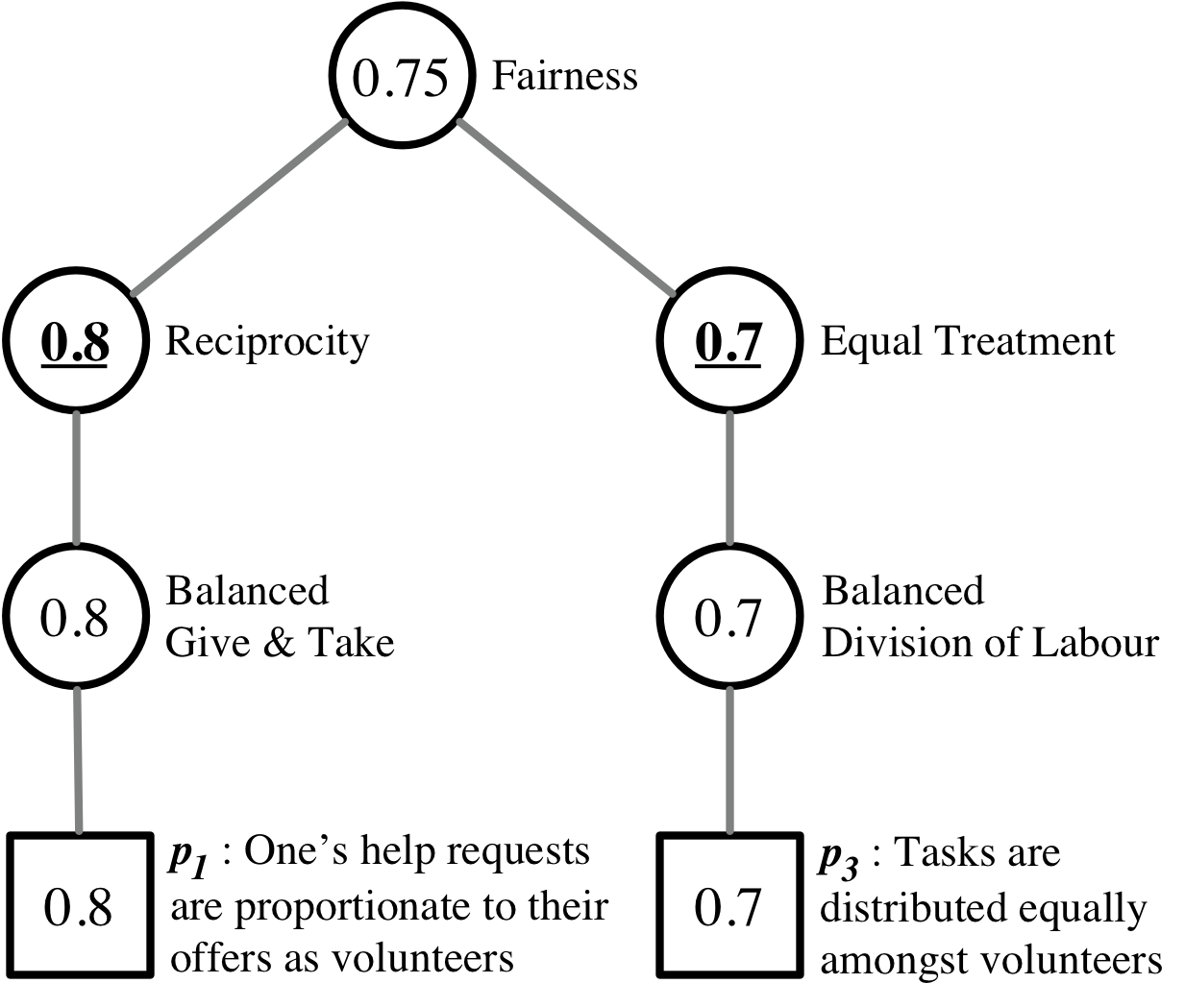}
         \caption{$\mathcal{V}_{c_{s}}^{user^{1}}$}
         \label{fig:user1}
     \end{subfigure}
     \hfill
     \begin{subfigure}[b]{.24\linewidth}
         \centering
         \includegraphics[width=\linewidth]{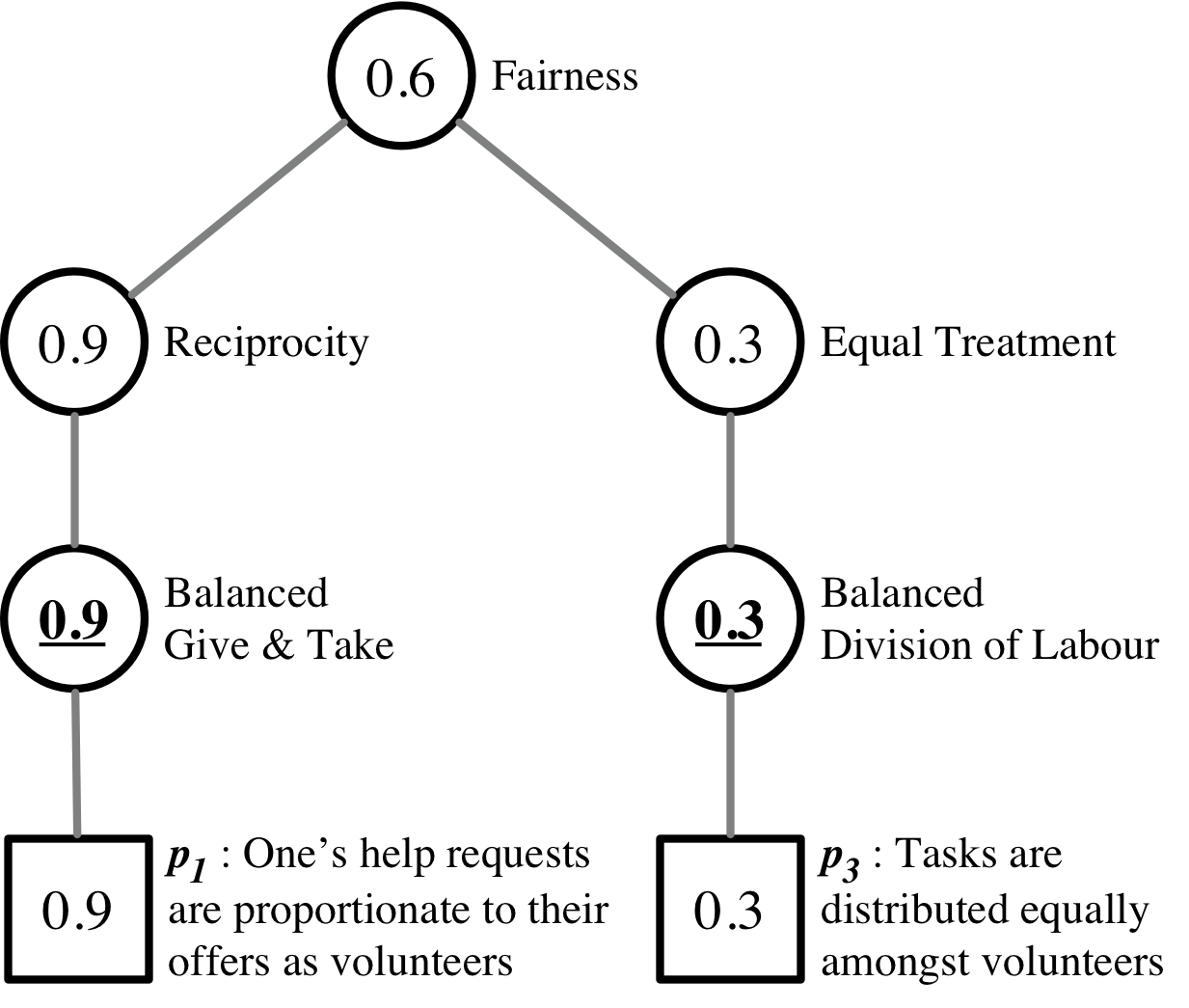}
         \caption{$\mathcal{V}_{c_{s}}^{user^{2}}$}
         \label{fig:user2}
     \end{subfigure}
     \hfill
     \begin{subfigure}[b]{.24\linewidth}
         \centering
         \includegraphics[width=\linewidth]{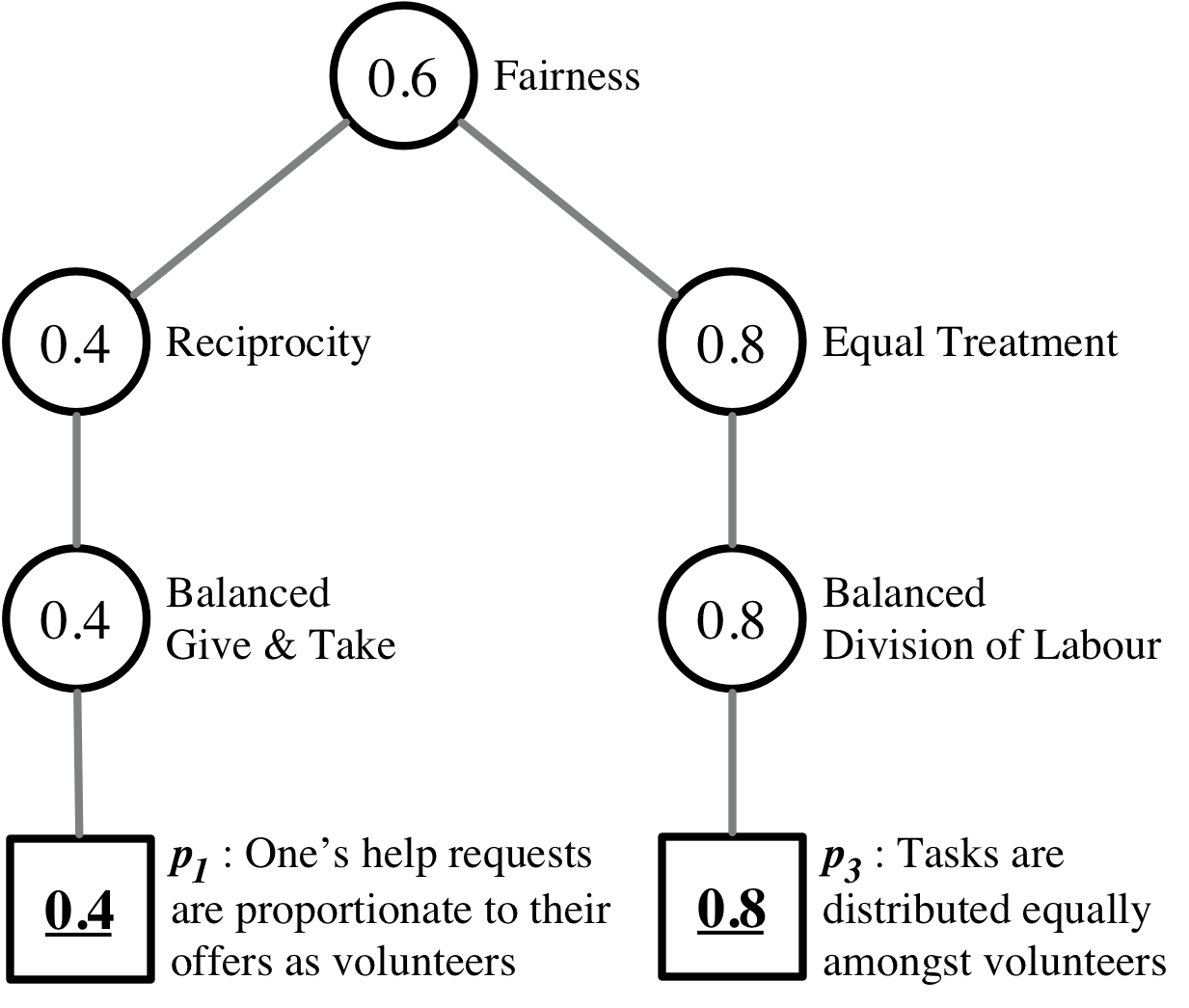}
         \caption{$\mathcal{V}_{c_{s}}^{user^{3}}$}
         \label{fig:user3}
     \end{subfigure}
     \hfill
     \begin{subfigure}[b]{.24\linewidth}
         \centering
         \includegraphics[width=\linewidth]{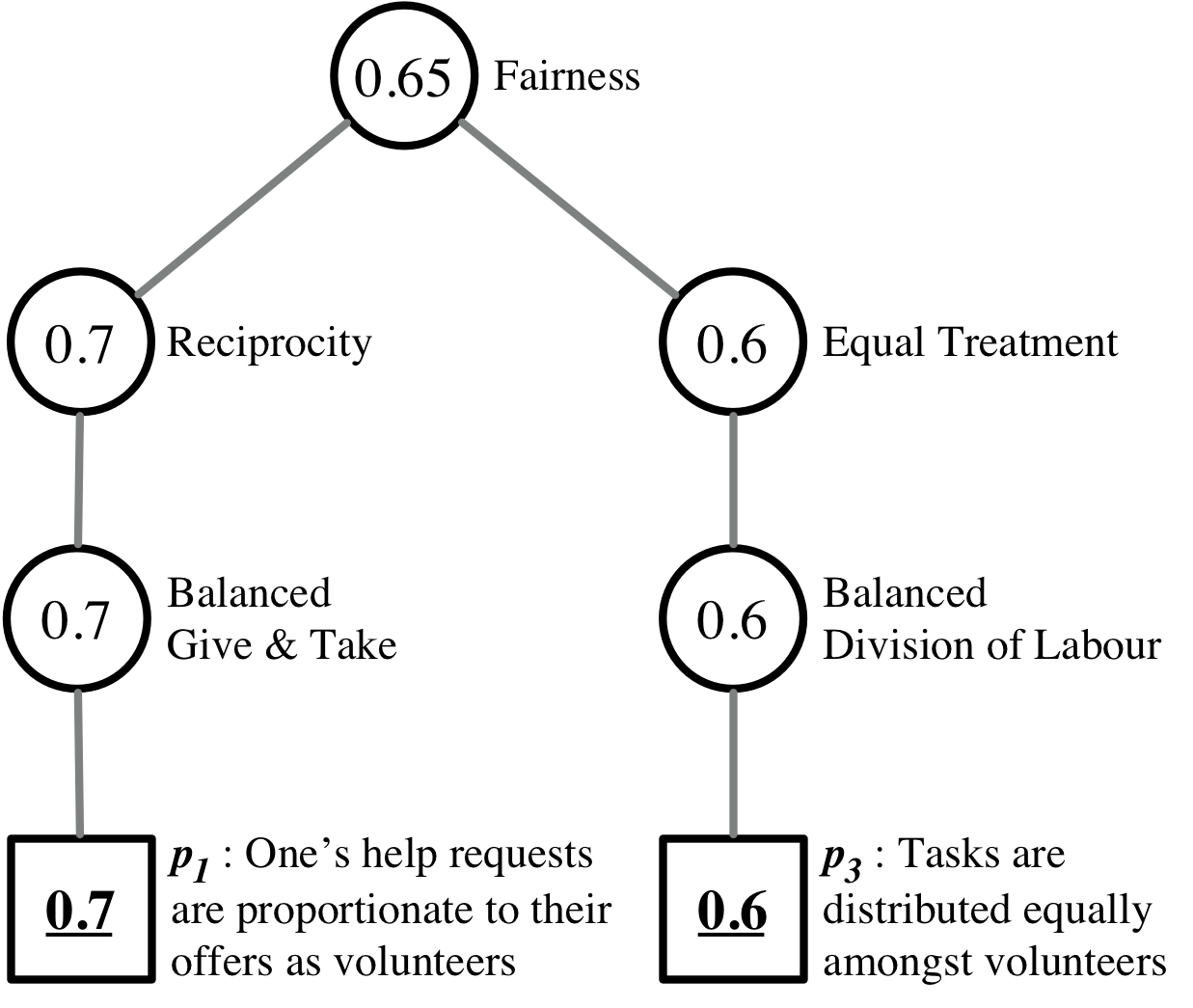}
         \caption{$\mathcal{V}_{c_{s}}^{collective}$}
         \label{fig:group}
     \end{subfigure}
    \caption{Individual and collective value taxonomies for fairness in uHelp's community of single mothers}
    \label{fig:individualVcollective}
\end{figure*}

\subsubsection{Implementation Choices}
The issue here is implementing mechanisms that take the value taxonomies of individuals and compute a taxonomy of these individuals as a collective. This is a complex task, and the domain dictates how collective values are specified. For example, the company developing an assistive robot may pre-define the value taxonomy governing the behaviour of that robot. In a hospital, an elected board of medics convene to agree on the hospital's values collectively. Yet. in other cases, such as the uHelp application, one can imagine the community of users coming together to vote on their values. In other words, the rules dictating whose view should be considered when specifying the value taxonomies of a collective are domain dependent.

Different mechanisms may be developed to construct collective values from individual ones.  
For example, negotiation and argumentation mechanisms can help individuals reach collective agreements on their values. Computational social choice may also be used to aggregate individual values into collective ones. We point the interested reader to ongoing work on this topic~\cite{Lera-LeriBSLR22}, where the aggregation takes into consideration various ethical principles, such as utilitarian (maximum utility) or egalitarian (maximum fairness).

\subsubsection{The Running uHelp Example}
This subsection highlights that both individuals and collectives may hold value taxonomies. For example, imagine three uHelp members from the community of single mothers who hold the taxonomies of Figures \ref{fig:user1}--\ref{fig:user3}. Of course, more members would exist in reality, each with their own taxonomy, but for simplicity, we show the taxonomies of three individuals only. Given these taxonomies, the social choice mechanism of~\cite{Lera-LeriBSLR22} can then be used to compute the importance of each property node for the collective. With those, a propagation mechanism (such as that of~\cite[Alg. 1]{ValuesFrameworkArXiV23}) can then be used to propagate those to the rest of the tree. The resulting taxonomy is shown in Figure~\ref{fig:group}.

\subsection{Why hold values? Value-alignment problem}\label{sec:why}
Values are one of the main motivators of behaviour~\cite{Schwartz2012AnOO,Rohan2000}. Assessing the alignment of behaviour with values has been the main objective of the work on values in AI to ensure value-aligned behaviour. The property-based nodes of a value taxonomy introduce the foundations for linking abstract value concepts to concrete computational constructs that can help formally assess the alignment of behaviour with those values. The value alignment of an entity's behaviour becomes the degree of satisfaction of the property-based value nodes of the relevant taxonomy that this behaviour brings about. In Definition~\ref{def:contextValueTaxonomy} we have seen how importance is assigned to different taxonomy nodes.\footnote{Value-alignment will be assessed within different contexts, which is discussed in Subsection~\ref{sec:context} to context-based value taxonomies.} The evaluation of value alignment should take into consideration these importance measures: the more important a property-based node is, the higher its satisfaction contributes to the value alignment of the behaviour being evaluated, and vice versa. The alignment of an entity $e$'s behaviour with a context-based value taxonomy $\mathcal{V}_{c}$ is then defined accordingly:\footnote{As before, context-based value taxonomies are expected to describe the values held by some entity. However, for the sake of simplifying notation, we drop the holder $x$ (and possibly $x$'s view of $y$'s values, if that was the case) and replace $\mathcal{V}^{x}_{c}$ with $\mathcal{V}_{c}$ (or $\mathcal{V}^{x>y}_{c}$ with $\mathcal{V}_{c}$). We also note that $x$ (or even $y$) does not necessarily have to be the same entity $e$ whose behaviour is being assessed. In other words, if $x=e'$, then this describes assessing how much $e$ is aligned with the values of $e'$.} 
\begin{equation}\label{eq:alignment}
    \mathcal{A}(e,\mathcal{V}_{c}) = \bigoplus_{p\in N_{\phi,c}}  f(sd(p,e),I_{c}(p))
\end{equation}
where $N_{\phi,c}$ represents the property nodes of the taxonomy $\mathcal{V}_{c}$, $sd(p,e)$ represents the degree of satisfaction of property $p$ with respect to the behaviour of entity $e$, and $I_{c}(p)$ represents the importance of the property-node $p$ within the context-based value taxonomy $\mathcal{V}_{c}$. The function $f$ is used to factor in the importance of property nodes when considering their degree of satisfaction, whereas $\bigoplus$ is used to aggregate the degree of satisfaction of all property nodes in $\mathcal{V}_{c}$ (with value importance factored in). 

\subsubsection{Implementation Choices}\label{sec:valueAlignment} 

One question is calculating the satisfaction of property nodes $sd(p,e)$. 
In other words, given an entity $e$, how do we assess to what degree the behaviour of $e$ results in the satisfaction of property $p$? This requires knowledge about how $e$ behaves, and different implementation approaches for specifying this knowledge can be followed. For example, suppose $e$ is a complex system of communicating entities. In that case, $e$'s model (usually specified via a process calculi) will describe its behaviour through a labelled transition system where the satisfaction of specific properties at different states~\cite{Stirling2001} can be evaluated. 
If $e$ is a normative system, then the norms can help map out the state diagram of the possible interaction outcome and evaluate the satisfaction of relevant properties accordingly~\cite{aagotnes2007logic,cranefield2011verifying,MontesOS22}. If $e$ was an agent with a BDI model, then BDI reasoning mechanisms can help assess the degree to which specific properties will be satisfied by $e$'s behaviour~\cite{rao1998decision}. In summary, a model of $e$ describing its behaviour is necessary to assess $sd(p,e)$. This issue has already been addressed in many fields, as illustrated above. To ensure our proposal is not limited to one modelling choice, we omit the choice of modelling $e$'s behaviour and assume the degree of satisfaction $sd(p,e)$ to be attainable. 

Returning to the alignment function $\mathcal{A}$, there are other implementation choices, such as the choice of the function $f$ that factors in the importance of a property node and the aggregation operator $\bigoplus$. In this paper, we propose a straightforward implementation that follows a weighted average approach: 
\begin{equation}\label{eq:alignmentImp}
    \mathcal{A}(e,\mathcal{V}_{c})= ( \displaystyle\sum_{p\in N_{\phi,c}}  I_{c}(p) \cdot sd(e,p)) \; / \; (|N_{\phi,c}|)
\end{equation}

Assuming the range of value importance $I$ to be $[-1,1]$, and degree of satisfaction $sd(e,p)$ a number with the range $[0,1]$, 
then the range of $\mathcal{A}$ becomes $[-1,1]$ where negative results describe the degree of misalignment (or an alignment with detested values) and positive results describe the degree of alignment with aspired values.

\subsubsection{The Running uHelp Example} 
Let us consider the context-based value taxonomy $\mathcal{V}_{c_{s}}'$ of Figure~\ref{fig:single-mothers-2} for a mutual aid community. The concrete definitions of $p_1$ and $p_3$ (property definitions \ref{eq:p1} and~\ref{eq:p3}) illustrate what it means, computationally, for the behaviour of some entity to be aligned with the value `fairness' in this context. Next, we illustrate how the exact degree of satisfaction of $p_1$ and $p_3$ can be computed according to these definitions. Equation~\ref{eq:p1Sat} formally states that the degree of satisfaction of $p_1$ is the actual ratio of requests to offers, normalised to fall into the range $[-1,1]$.
\begin{equation}\label{eq:p1Sat}
    sd(e,p_1)= 
    \begin{cases}
        (R -1) \; / \; ((\max R) - 1)
            &  \text{, if } R > 1 \\[1em]
        R -1
            & \text{, otherwise}
    \end{cases}
\end{equation}
where $R=\# requests/\# \mathit{offers}$ represents the ratio of requests to offers, and $\max R$ is the maximum possible value for $R$. While the range of $R$ is $[0,\infty)$, a maximum value must be selected for our equations. We argue that $\max R$ is domain dependent and should be selected for each scenario.  
Equation~\ref{eq:p1Sat} states that the degree of satisfaction is computed by normalising the ratio $R$ of requests to offers to the range $[-1,1]$. The degree of satisfaction of $p_1$ depends on how far is the ratio $R$ from $1$. The larger it is with respect to $1$, the higher the degree of satisfaction. The closer it is to $0$, the greater the degree of dissatisfaction.

Next, Equation~\ref{eq:p3Sat} defines the satisfaction of property $p_3$ similarly by formally stating that the degree of satisfaction of $p_3$ is the actual difference between the uniform distribution $U$ and the distribution of tasks over volunteers $D$, normalised to the range [-1,1].
\begin{equation}\label{eq:p3Sat}
    sd(e,p_3)= 
    \begin{cases}
        1 - (\Delta \;/\; \epsilon) 
            & \text{, if } \Delta < \epsilon \\[1em]
        (- (\Delta - \epsilon)) \; / \; ((\max \Delta) - \epsilon) 
            & \text{, otherwise}
    \end{cases}
\end{equation}
where $\Delta=$ \emph{difference} $(D,U)$ represents the difference between the distribution of tasks over volunteers ($D$) and the uniform distribution ($U$), and $\max \Delta$ is the maximum possible value for $\Delta$. The range of $\Delta$ is $[0,\infty)$, but a maximum value must be selected for our equations. Again, we argue that $\max \Delta$ is domain dependent, and must be chosen for each scenario. 
Equation~\ref{eq:p3Sat} states that the degree of satisfaction of $p_3$ depends on how far is the difference $\Delta$ from $\epsilon$. The larger it is with respect to $\epsilon$, the higher the degree of dissatisfaction. The closer it is to $0$, the greater the degree of satisfaction.

Now, say a mutual aid community $e$ provides incentives that motivate people to volunteer, %more, 
and has norms that ensure tasks are spread as equally as possible over volunteers. 
Say the regimented norms result in a  high degree of satisfaction for $p_3$, whereas the incentives result in a mediocre degree of satisfaction for $p_1$:
$$\begin{array}{ccc}
sd(e,p_1) = 0.5 & ; &
sd(e,p_3) = 0.9
\end{array}$$
And say the importance of $p_1$ is set to be twice that of $p_2$: %, as follows: 
$$\begin{array}{ccc}
I_{c}(p_1) = 1 & ; &
I_{c}(p_3) = 0.5
\end{array}$$

Following Equation~\ref{eq:alignmentImp}, the alignment of the mutual aid community $e$ with its understanding of fairness $\mathcal{V}_{c_{s}}'$ (Figure~\ref{fig:single-mothers-2}) becomes:
\begin{align*}
    \mathcal{A}(e,\mathcal{V}_{c}) & %= \frac{(I_{c}(p_1)\cdot sd(e,p_1)) + (I_{c}(p_3) \cdot sd(e,p_3))}{2} 
    = 0.475
\end{align*}

%%%%%%%%%%%%%%%%%%%%%%%%%%%%%%%%%%%%%%%%%%%%%%%%%%%%%%%%%%%%%%%%%%%%%%%%

\section{Contributions and Further Work}\label{sec:StrengthLimit}
The novel contribution is a formal and computational model of human values that lays the foundations for algorithmic reasoning over them, which is strongly aligned with existing research from social psychology. 
Schwartz and Bilsky's five features that are recurrently mentioned in the literature to define values state that values ``(1) are concepts or beliefs, (2) pertain to desirable end states or behaviours, (3) transcend specific situations, (4) guide selection or evaluation of behaviour and events, and (5) are ordered by relative importance''~\cite{Schwartz1987TowardAU}.

These features are shared by many social psychologists and social scientists~\cite{Rohan2000}, aligning with our value taxonomy proposal as follows. Values are abstract concepts (feature 1), specified through `labels' like fairness, equality, etc. They are defined through desirable end states (feature 2), implemented via property nodes. The whole work on values is to guide behaviour (feature 4): our value-alignment mechanism assesses to what extent behaviour is aligned with selected values. Value importance is integral to our approach (feature 5), where node importance is critical for computing value alignment. While feature 3 states that values transcend specific situations, we argue that although value taxonomies do not change frequently, they do evolve. Here, we are more aligned with the work in value-sensitive design~\cite{vandePoel2018}.  

One major challenge is designing and constructing value taxonomies. The human values we want to specify and embed in the decision-making processes need to come from diverse human stakeholders, including users, designers, owners, and others directly or indirectly affected.  One approach involves stakeholders explicitly detailing their values in a way that can be directly mapped to a formal model such as ours; another involves AI learning stakeholder values from their interactions. 
The first approach requires significant effort from humans, while the second is prone to errors in the learning mechanism. While impressive value learning~\cite{10.5555/3463952.3464048} and value aggregation~\cite{Lera-LeriBSLR22} mechanisms are being proposed, they are not error-free, and they do not deal with the complexity of value taxonomies. Introducing these taxonomies introduces new challenges, such as learning the importance of values, the relations between value nodes, the property nodes for some value concepts, and designing mechanisms to aggregate value taxonomies. 

The second issue concerns the modelling process, necessitating representation choices that may bring limitations.
We have introduced guiding principles that support our modelling decisions and aimed to consider theoretical notions that clarify implementation choices. Whilst we deliberately leave the choice of representation and implementation open for system and research development, we have made implementation choices in our running example. 
\section{Conclusions}\label{sec:conclusion}
We have contributed to the urgent challenge of building value-aligned AI by proposing a conceptually intuitive foundational model for human values.
It allows for future computational reasoning and opens up opportunities to evidence how AI systems are \emph{provably} aligned with human values. 
The approach is grounded in social psychology, subsumes existing AI research concepts, and is formal, making it a coherent and intuitive starting point for future interdisciplinary research investigation. 
To our knowledge, this is the first proposal for the formal representation of human values and moves beyond the state-of-the-art ---which to date has defined values through labels~\cite{10.5555/3463952.3464048,SerramiaLR20} or goals~\cite{abs-2110-09240,MontesS21}--- by explicitly introducing notions of value importance, semantics, and relations.

\begin{acks}
This work has been supported by the EU-funded VALAWAI (\#~101070930) project and the Spanish-funded VAE (\#~TED2021-131295B-C31) and Rhymas (\#~PID2020-113594RB-100) projects.
\end{acks}

%%%%%%%%%%%%%%%%%%%%%%%%%%%%%%%%%%%%%%%%%%%%%%%%%%%%%%%%%%%%%%%%%%%%%%%%

%%% The next two lines define, first, the bibliography style to be 
%%% applied, and, second, the bibliography file to be used.

\bibliographystyle{ACM-Reference-Format} 
\bibliography{references}

%%%%%%%%%%%%%%%%%%%%%%%%%%%%%%%%%%%%%%%%%%%%%%%%%%%%%%%%%%%%%%%%%%%%%%%%

\end{document}